\documentclass[letterpaper, 10pt, conference]{ieeeconf/ieeeconf}    

\makeatletter
\let\NAT@parse\undefined
\makeatother

\usepackage[numbers,sectionbib,sort&compress]{natbib}

\usepackage{bm}
\usepackage{gensymb}
\usepackage{graphicx}
\usepackage{amsmath}
\usepackage{algorithm}
\usepackage[noend]{algpseudocode}
\usepackage{subcaption}
\usepackage{amsfonts}
\usepackage{siunitx}
\usepackage{booktabs}
\usepackage[font=small]{caption}
\usepackage[export]{adjustbox}
\usepackage{sidecap} \sidecaptionvpos{figure}{c}
\DeclareMathOperator*{\argmax}{\arg\!\max}

\DeclareMathOperator{\Tr}{Tr}

\usepackage{hyperref}
\hypersetup{colorlinks,breaklinks,
linkcolor=[rgb]{0.5,0.,0.},
citecolor=[rgb]{0.000,0.427,0.173},
urlcolor=[rgb]{0.031,0.318,0.612}}

\usepackage[printonlyused,withpage,nolist,nohyperlinks]{acronym}
\begin{acronym}

\acro{2D}{two-dimensional}

\acro{3D}{three-dimensional}

\acro{AHRS}{attitude and heading reference system}

\acro{AUV}{autonomous underwater vehicle}

\acro{CPP}{Chinese Postman Problem}

\acro{DoF}{degree of freedom}
\acrodefplural{DoF}[DoFs]{degrees of freedom}

\acro{DVL}{Doppler velocity log}

\acro{FSM}{finite state machine}

\acro{IMU}{inertial measurement unit}

\acro{LBL}{Long Baseline}

\acro{MCM}{mine countermeasures}

\acro{MDP}{Markov decision process}
\acrodefplural{MDP}[MDPs]{Markov decision processes}

\acro{POMDP}{partially observable Markov decision process}
\acrodefplural{POMDP}[POMDPs]{partially observable Markov decision processes}

\acro{PRM}{Probabilistic Roadmap}
\acrodefplural{PRM}[PRM]{Probabilistic Roadmaps}

\acro{ROI}{region of interest}
\acrodefplural{ROI}[ROIs]{regions of interest}

\acro{ROS}{Robot Operating System}

\acro{ROV}{remotely operated vehicle}

\acro{RRT}{Rapidly-exploring Random Tree}
\acrodefplural{RRT}[RRTs]{Rapidly-exploring Random Trees}

\acro{SLAM}{simultaneous localization and mapping}

\acro{SSE}{sum of squared errors}

\acro{STOMP}{Stochastic Trajectory Optimization for Motion Planning}

\acro{TRN}{Terrain-Relative Navigation}

\acro{UAV}{unmanned aerial vehicle}

\acro{USBL}{Ultra-Short Baseline}

\acro{IPP}{informative path planning}

\acro{FoV}{field of view}
\acrodefplural{FoV}[FoVs]{fields of view}

\acro{CDF}{cumulative distribution function}

\acro{ML}{maximum likelihood}

\acro{RMSE}{Root Mean Squared Error}
\acro{MLL}{Mean Log Loss}

\acro{GP}{Gaussian Process}
\acrodefplural{GP}[GPs]{Gaussian processes}

\acro{KF}{Kalman Filter}

\acro{IP}{Interior Point}
\acro{BO}{Bayesian Optimization}

\end{acronym}

\providecommand{\figref}[1]{\mbox{Fig.~\ref{#1}}}
\providecommand{\eqnref}[1]{\mbox{Eq.~\ref{#1}}}
\providecommand{\tabref}[1]{\mbox{Table~\ref{#1}}}
\renewcommand{\algref}[1]{\mbox{Alg.~\ref{#1}}}
\providecommand{\secref}[1]{\mbox{Section \ref{#1}}}

\IEEEoverridecommandlockouts                           
\title{\LARGE \bf
Multiresolution Mapping and Informative Path Planning \\ for UAV-based Terrain Monitoring
}

\author{$\text{Marija Popovi\'{c}}$, $\text{Teresa Vidal-Calleja}$, $\text{Gregory Hitz}$, 
$\text{Inkyu Sa}$, $\text{Roland Siegwart}$, and $\text{Juan Nieto}$
\thanks{M. Popovi\'{c}, G. Hitz, J. Nieto, I. Sa, and R. Siegwart are with the Autonomous Systems Lab., ETH Z\"{u}rich, Z\"{u}rich, Switzerland.
T. Vidal-Calleja is with the Centre for Autonomous Systems, University of Technology, Sydney, Australia.
Corresponding author: \texttt{mpopovic@ethz.ch}.}%
}

\begin{document}

\maketitle
\thispagestyle{empty}
\pagestyle{empty}

\begin{abstract}
Unmanned aerial vehicles (UAVs) can offer timely and cost-effective delivery of high-quality sensing data.
However, deciding when and where to take measurements in complex environments remains an open challenge.
To address this issue,
we introduce a new multiresolution mapping approach for informative path planning
in terrain monitoring using UAVs.
Our strategy exploits the spatial correlation encoded in a Gaussian Process model
as a prior for Bayesian data fusion with probabilistic sensors.
This allows us to incorporate altitude-dependent sensor models for aerial imaging
and perform constant-time measurement updates.
The resulting maps are used to plan information-rich trajectories in continuous 3-D space
through a combination of grid search and evolutionary optimization.
We evaluate our framework on the application of agricultural biomass monitoring.
Extensive simulations show that our planner performs better than existing methods,
with mean error reductions of up to 45\% compared to traditional ``lawnmower'' coverage.
We demonstrate proof of concept using a multirotor \acused{UAV} to map color in different environments.
\end{abstract}

\section{INTRODUCTION} \label{S:introduction}

Environmental monitoring provides valuable scientific data helping us better understand the Earth and its evolution.
However, typically targeted natural phenomena exhibit complicated patterns with high spatio-temporal variability.
Despite this complexity, much scientific data is still acquired using portable or static sensors in arduous and potentially dangerous campaigns.
In both the marine~\citep{Hitz2014} and terrestrial~\citep{Popovic2017, Marchant2014} domains, mobile robots offer a cost-efficient, flexible alternative
enabling data-gathering at unprecedented levels of resolution and autonomy~\cite{Dunbabin2012}.
These applications have opened new challenges in mapping and path planning for large-scale monitoring given platform-specific constraints.

This paper focuses on mapping strategies for \ac{IPP} in \ac{UAV}-based agricultural monitoring.
Our motivation is to increase the efficiency of data collection on farms by using an on-board camera to quickly find areas requiring treatment.
As well as enabling coverage, this workflow provides detailed data for decisions to reduce chemical usage and optimize yield~\citep{Liebisch2017,Cardina1997}.
A key challenge in this set-up is fusing visual information received from different altitudes into a single probabilistic map.
Using the map, the \ac{IPP} unit must identify most useful future measurement sites; trading off between image resolution and \ac{FoV}
while accounting for limited battery and computational resources.

\begin{figure}[!t]
\centering
  \begin{subfigure}[]{0.13\textwidth}
  \includegraphics[width=\textwidth]{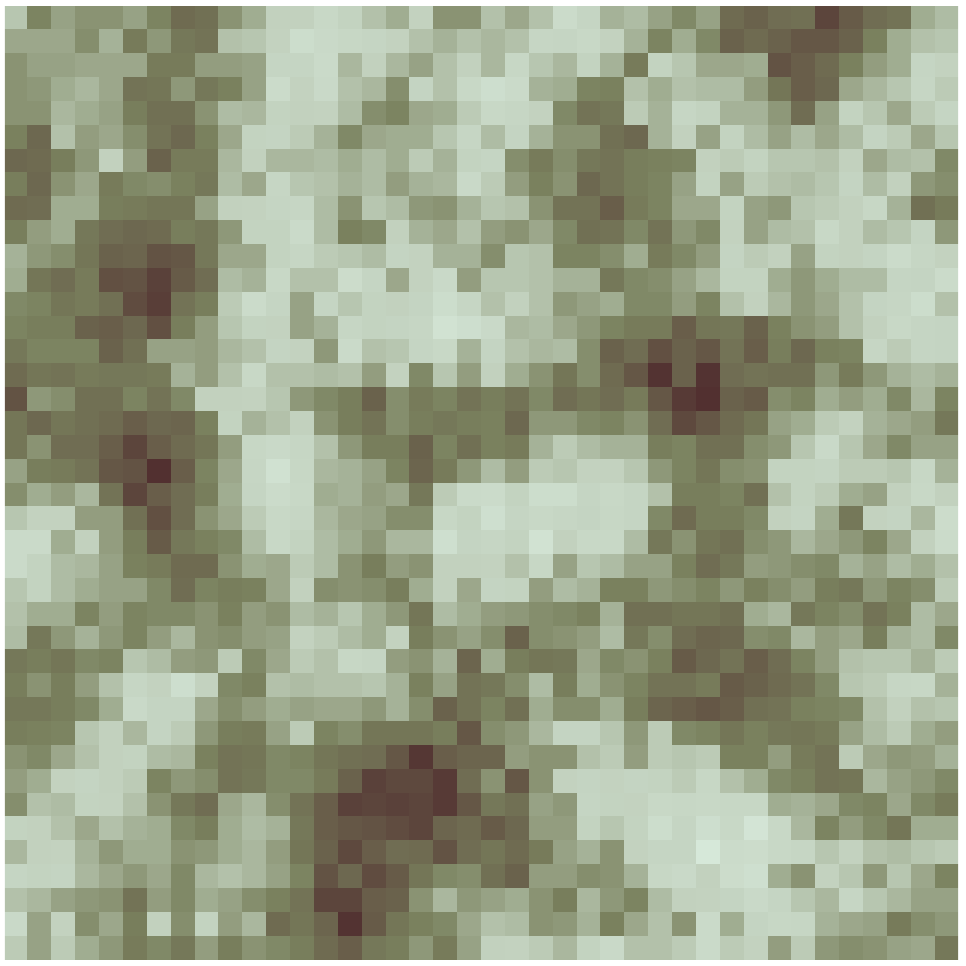}
  \caption{}
  \end{subfigure}\hfill%
  \begin{subfigure}[]{0.13\textwidth}
  \includegraphics[width=\textwidth]{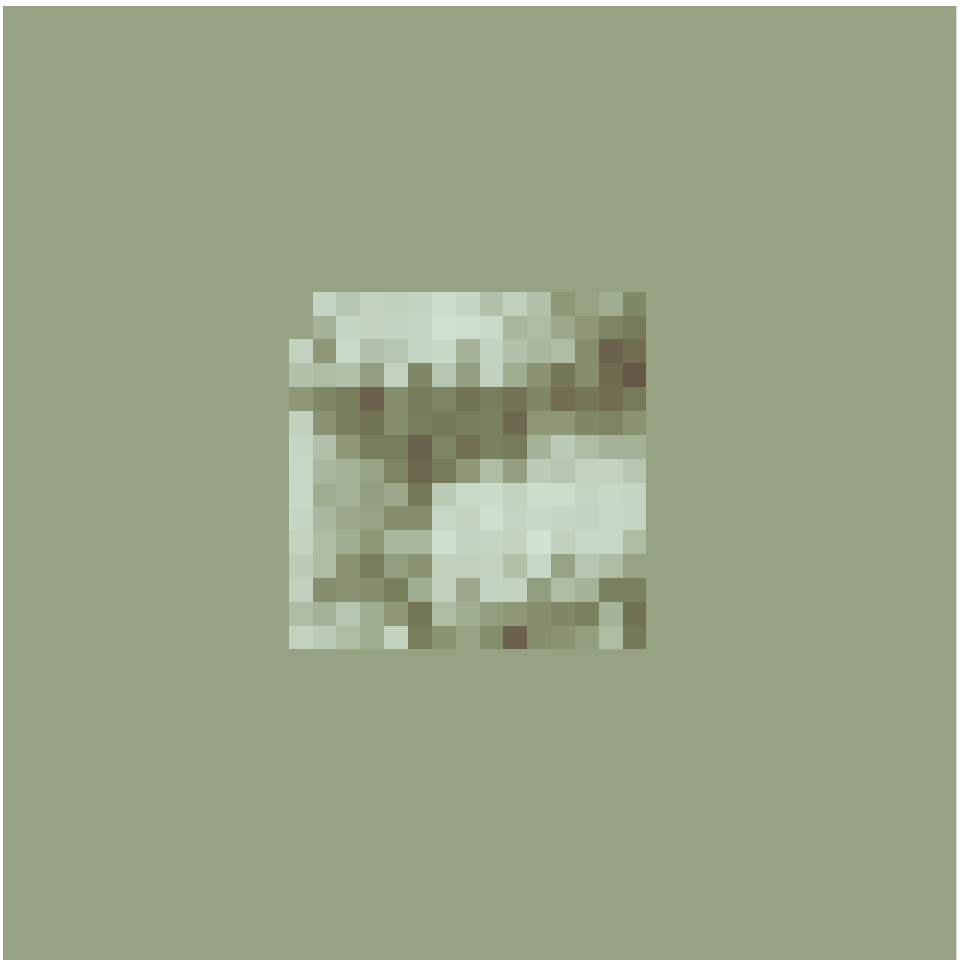}
  \caption{}
  \end{subfigure}\hfill%
  \begin{subfigure}[]{0.13\textwidth}
  \includegraphics[width=\textwidth]{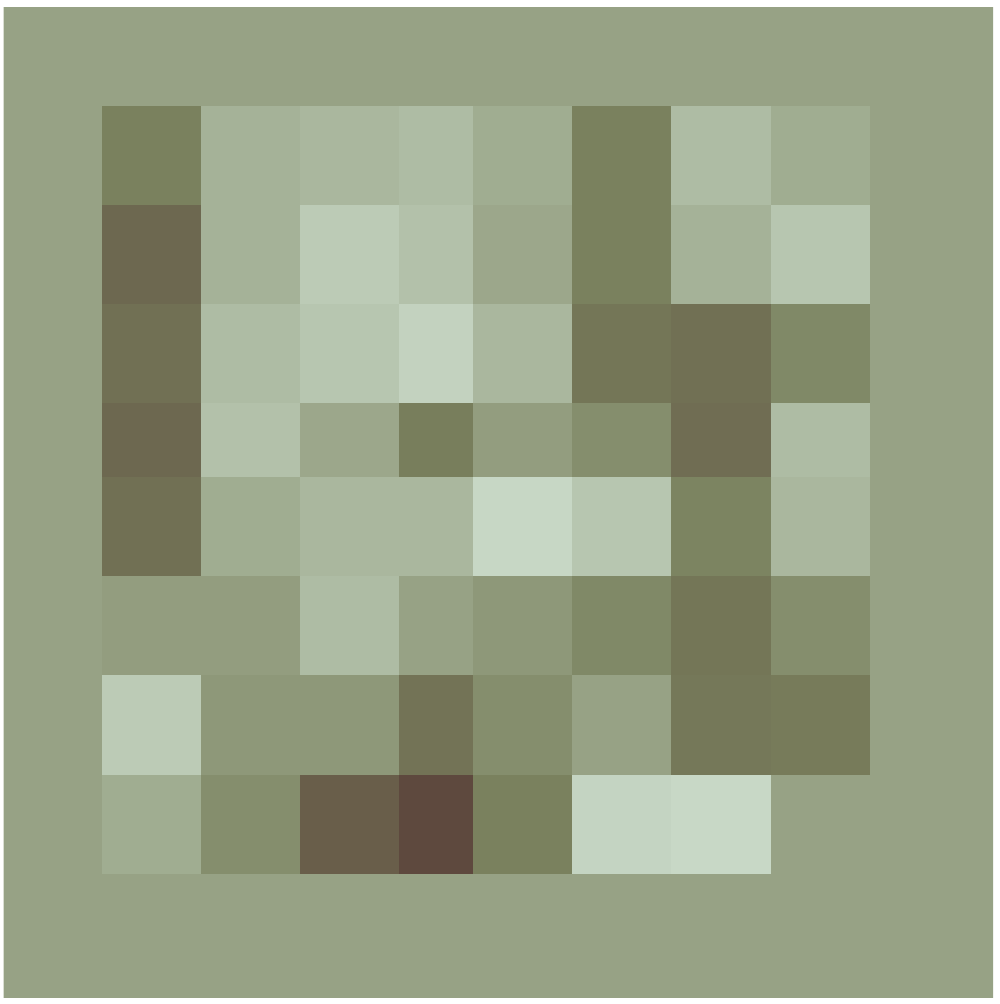}
  \caption{}
  \end{subfigure}
    
  \vspace{0.2mm}
  \begin{subfigure}[]{0.14\textwidth}
  \hspace{11mm}
  \includegraphics[height=\textwidth]{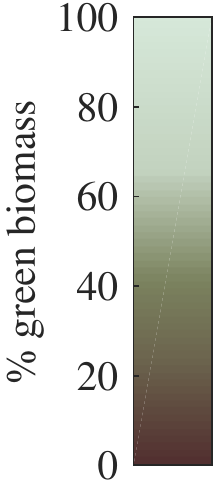}
  \vspace{5mm}
  \end{subfigure}\hfill%
  \begin{subfigure}[]{0.13\textwidth}
  \hspace{-2mm}
  \includegraphics[width=\textwidth]{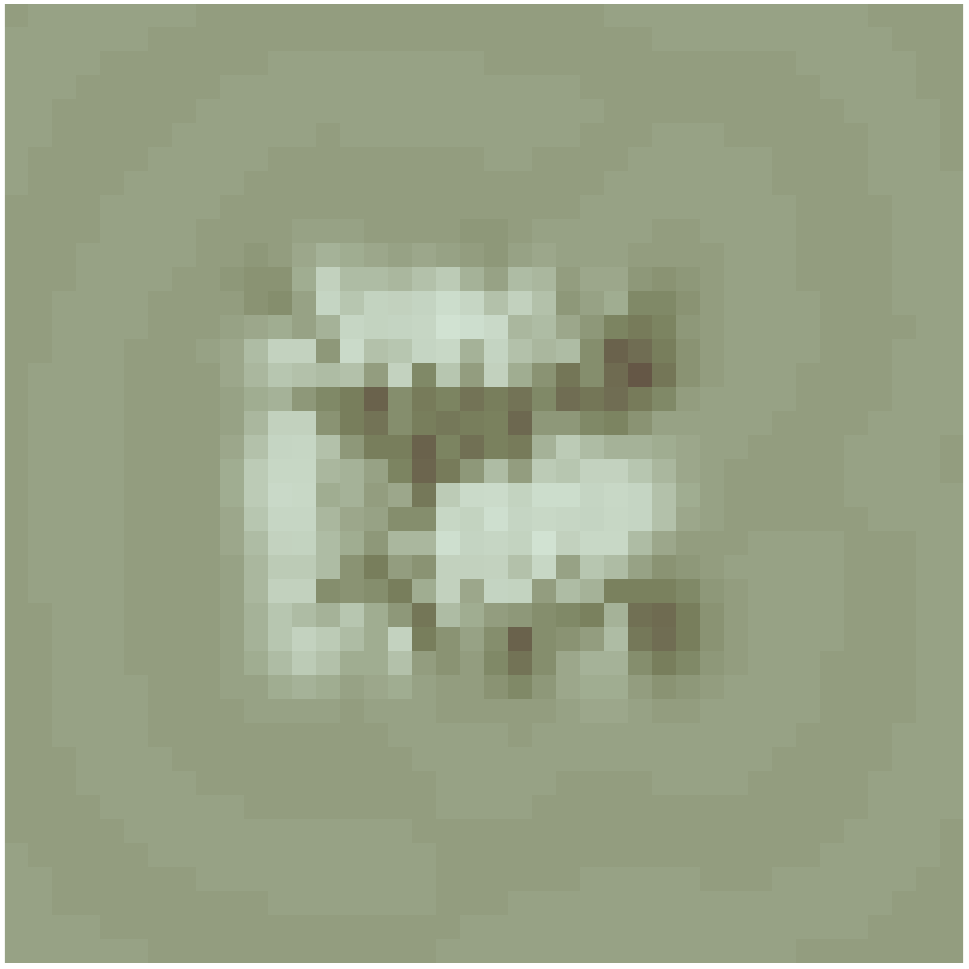}
  \caption{}
  \end{subfigure}\hfill%
  \begin{subfigure}[]{0.13\textwidth}
  \includegraphics[width=\textwidth]{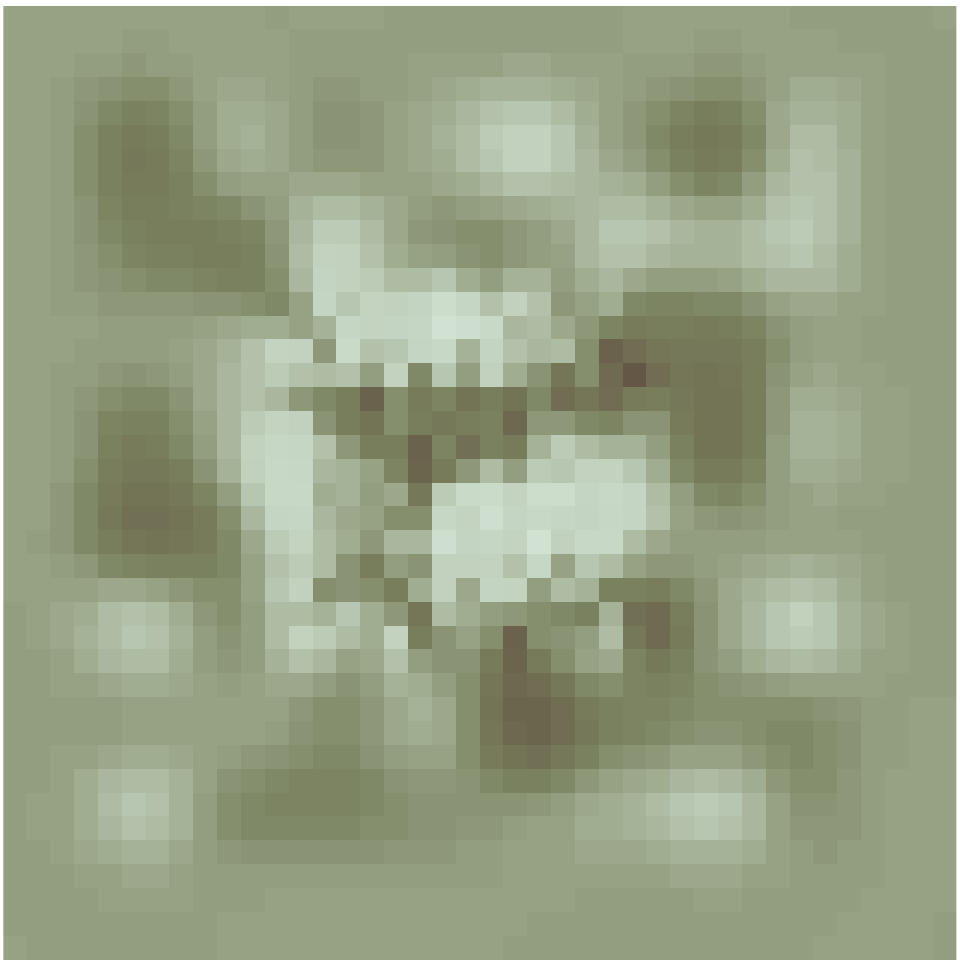}
  \caption{}
  \end{subfigure}
  
   \caption{Overview of our proposed mapping strategy.
   A ground truth biomass distribution is shown in (a).
   (b) and (c) depict variable-resolution measurements taken from $9$\,m and $20$\,m altitudes.
   (d) and (e) illustrate maps resulting from fusing the data sequentially.
   The variable diffusion effects show that our method can handle uncertain multiresolution measurements
   and capture spatial correlations.
   }\label{F:teaser}
\end{figure}

We address these problems by presenting a new multiresolution mapping approach for \ac{IPP}.
In this paper, we build upon methods established in our prior work~\citep{Popovic2017}.
As observing continuous (e.g. green biomass cover), rather than binary (e.g. weed occupancy), variables is of interest in agriculture,
our framework is adapted to meet these requirements.
We use \acp{GP} as a natural way of encoding spatial correlations common in biomass distributions~\citep{Cardina1997}.
A key aspect of our method is the use of Bayesian fusion for sequential map updates based on the \ac{GP} prior.
This enables us to perform constant-time measurements with an altitude-dependent sensor model (\figref{F:teaser}).
Using this map, our planning scheme applies an evolutionary technique to optimize trajectories initialized by a 3-D grid search.
We introduce a parametrization to cater for realistic sensor dynamics and a strategy to adaptively focus on areas with high infestation probability.
The contributions of this work are:
\begin{enumerate}

 \item A new recursive mapping approach which:
  \begin{itemize}
   \item uses \acp{GP} as priors for multiresolution data fusion,
    \item enables constant-time map updates,
     \item supplies uncertainty information for \ac{IPP}.
  \end{itemize}
  \item The evaluation of our framework against state-of-the-art planners
  and alternative optimization methods.
  \item Proof of concept through autonomously executed tests.
\end{enumerate}

While our framework is motivated by an agricultural application, we note that it can be used in any scalar field mapping scenario,
e.g. pipe thickness~\citep{Vidal-Calleja2014}, spatial occupancy~\cite{Charrow2015}, gas concentration~\citep{Marchant2014}, elevation~\citep{Colomina2014}, etc.

\section{RELATED WORK} \label{S:related_work}
A large body of prior work has studied \ac{IPP} for mapping and exploration.
Research in this field can be divided into two main areas:
methods of environmental modelling~\citep{Rasmussen2006,Vasudevan2009,OCallaghan2012,Vidal-Calleja2014,Sun2015}
and algorithms for efficient data acquisition~\citep{Binney2013,Hitz2014,Hollinger2014,Marchant2014}.

\acp{GP} are a popular Bayesian technique for modeling correlations in spatio-temporal phenomena~\citep{Rasmussen2006}.
For \ac{IPP}, they have been applied in various scenarios~\citep{Marchant2014,Hollinger2014,Hitz2014,Vivaldini2016}
to collect data accounting for map structure and uncertainty.
This framework permits using both
covariance functions to express complex dependencies~\citep{Singh2010} and approximations to handle large datasets~\citep{Rasmussen2006}.
Our work follows these lines by using \acp{GP} to create terrain maps~\citep{Vasudevan2009,Vidal-Calleja2014}
of continuous scalar fields.
In our application, we capture biomass level as a percentage with an associated uncertainty,
allowing us to quantify the utility of potential measurement sites.

A relatively unexplored aspect in research is building such models using aerial imagery.
Like~\citet{Vivaldini2016}, we study this set-up with a probabilistic sensor model.
The main issue with applying \acp{GP} here directly~\citep{Hitz2014,Vivaldini2016,Marchant2014,Hollinger2014}
is the computational load arising as dense imagery data accumulate over time.
We tackle this issue by using the spatial correlation of the \ac{GP} model as a prior for Bayesian data fusion~\citep{Vidal-Calleja2014},
thus procuring quicker map updates while accomodating multiple sensor modalities.
The efficiency of this approach can further be increased by using submaps~\citep{Sun2015}.

\citet{Krause2008} examine placing static sensors using a \ac{GP} model for maximum information gain.
\ac{IPP} extends this task by connecting measurement sites given robot mobility constraints.
Broadly, we distinguish between (i)~discrete and (ii)~continuous \ac{IPP} methods.
Whereas discrete algorithms, e.g. \citep{Binney2013}, operate on pre-defined grids,
continuous methods such as ours involve sampling strategies~\citep{Hollinger2014} or splines~\citep{Charrow2015,Hitz2014,Marchant2014},
and offer better scalability.
We follow the latter approaches in defining smooth polynomial trajectories for the \ac{UAV}~\citep{Richter2013},
which are optimized globally for an informative objective as in our prior work~\citep{Popovic2017}.

\ac{IPP} schemes are also classified based on their properties of adaptivity.
Unlike non-adaptive approaches~\citep{Binney2013,Hollinger2014,Vivaldini2016},
adaptive methods~\cite{Hitz2014,Sadat2015} allow plans to change as information is collected.
\citet{Sadat2015} devise an adaptive coverage planner for \acp{UAV} in applications similar to ours.
Their strategy, however, assumes discrete viewpoints and does not support probabilistic data acquisition.
In contrast, our work uses uncertain sensor models for data fusion
and incrementally replans continuous 3-D trajectories.

\section{PROBLEM STATEMENT} \label{S:problem_statement}
We define the general \ac{IPP} problem as follows. We seek a continuous trajectory $\psi$
in the space of all trajectories $\Psi$ for maximum gain in some information-theoretic measure:
\begin{equation}
\begin{aligned}
  \psi^* ={}& \underset{\psi \in \Psi}{\argmax}
\frac{I[\textsc{measure}(\psi)]}{\textsc{time}(\psi)}\textit{,} \\
 & \text{s.t. } \textsc{time}(\psi) \leq B \textit{,}
 \label{E:ipp_problem}
\end{aligned}
\end{equation}
where $B$ denotes a time budget and $I[$\textperiodcentered$]$ defines the utility function 
quantifying the informative objective. The function \textsc{measure(\textperiodcentered)} obtains discrete measurements along the
trajectory $\psi$ and \textsc{time(\textperiodcentered)} provides the corresponding
travel time.

In contrast to our previous work~\citep{Popovic2017}, the problem is formulated in the space of trajectories $\Psi$.
This allows us to express paths as functions of time and obtain measurements along them with a constant frequency,
thus reflecting the triggering operations found in practical devices.

\section{MAPPING APPROACH} \label{S:mapping_approach}
This section introduces our new mapping strategy as the basis of our \ac{IPP} framework.
In brief, a \ac{GP} is used to initialize a recursive filtering procedure,
thus replacing the computational burden of applying \acp{GP} directly with constant processing time in the number of measurements.
We first describe our method of creating prior maps
before detailing a Bayesian approach to fusing data from probabilistic sensors.

\subsection{Gaussian Processes} \label{s:gps}
We use a \ac{GP} to model spatial correlations in a probabilistic and non-parametric manner~\citep{Rasmussen2006}.
The target variable for mapping is assumed to be a continuous function in 2-D space: $\zeta: \mathcal{E} \rightarrow \mathbb{R}$,
where $\mathcal{E} \subset \mathbb{R}^2$ is the environment where measurements are taken. 
Using the GP, a Gaussian correlated prior is placed over the function space,
which is fully characterized by the mean $\bm{\mu} = E[\bm{\zeta}]$ and covariance $P = E[(\bm{\zeta}-\bm{\mu})(\bm{\zeta}^\top-\bm{\mu}^\top)]$ as $\bm{\zeta} \sim \mathcal{GP}(\bm{\mu},P)$.

Given a pre-trained kernel $K(X,X)$ for a fixed-size environment discretized at a certain resolution with $n$ locations $X \subset \mathcal{E}$,
we first specify a finite set of new prediction points $X^* \subset \mathcal{E}$ at which the prior map is to be inferred.
For unknown environments\footnote{Note that, for known environments, the \ac{GP} can be trained from available data and inferred at the same or different resolutions.},
as in our set-up, the values at $\textbf{x}_i \in X$ are initialized uniformly with a constant prior mean.
The covariance, however, is calculated using the classic GP regression equation~\citep{Reece2013}:
\begin{align}
 P ={}& K(X^*,X^*) - K(X^*,X)[K(X,X) + \sigma_n^2I]^{-1} \times &\notag\\
 &K(X^*,X)^\top \textit{,} \label{E:gp_cov}
\end{align}
where $P$ is the posterior covariance, $\sigma_n^2$ is a hyperparameter representing noise variance,
and $K(X^*,X)$ denotes cross-correlation terms between the predicted and initial locations.
%

To describe vegetation, we propose using the isotropic Mat\'ern 3/2 kernel function common in geostatistical analysis.
It is defined as~\citep{Rasmussen2006}:
\begin{align}
  k_{Mat3}(\textbf{x},\textbf{x}^*) =
  \sigma^2_f(1+ \frac{\sqrt{3}d}{l})\exp{(-\frac{\sqrt{3}d}{l})} \textit{,} \label{E:matern_kernel}
\end{align}
where $d$ is the Euclidean distance between inputs $\textbf{x}$ and $\textbf{x}^*$,
and $l$ and $\sigma_f^2$ are hyperparameters representing the lengthscale and signal variance, respectively.

The resulting set of fixed hyperparameters $\bm{\theta} = \{\sigma_n^2, \sigma_f^2, l\}$ controls relations within the \ac{GP}.
These values can be optimized using various methods~\citep{Rasmussen2006}
to match the properties of $\zeta$ by training on multiple maps at the required resolution.

Once the correlated prior map $p(\bm{\zeta}|X)$  is obtained, independent noisy measurements of variable resolution
are fused as described in the following section.

\subsection{Sequential Data Fusion} \label{s:data_fusion}
A key component of our framework is our map update procedure based on recursive filtering.
Given a uniform mean and the spatial correlations captured with Equation~\eqref{E:gp_cov},
the map $p(\bm{\zeta}|X) \sim \mathcal{GP}(\bm{\mu}^-, P^-)$ is used as a prior for fusing new sensor measurements.

Let $\bm{z} = [z_1, \ldots, z_m]^\top$ denote new $m$ independent measurements received at points $[\textbf{x}_1, \ldots, \textbf{x}_m]^\top \subset X$
modelled assuming a Gaussian sensor as $p(z_i|\zeta_i, \textbf{x}_i) = \mathcal{N}(\mu_{s,i}, \sigma_{s,i})$, as described in \secref{s:sensor_model}.
To fuse the measurements $\bm{z}$ with the prior map $p(\bm{\zeta}|X)$,
we use the maximum \textit{a posteriori} estimator, formulated as:
\begin{align}
 \argmax_{\bm{\zeta}} p(\bm{\zeta}|\bm{z},X)
\end{align}
To compute the posterior density $p(\bm{\zeta}|\bm{z},X) \propto p(\bm{z}|\bm{\zeta},X) \times p(\bm{\zeta}|X) \sim \mathcal{GP}(\bm{\mu}^+, P^+)$,
we directly apply the \ac{KF} update equations~\citep{Reece2013}:
\begin{align}
 \bm{\mu}^+ ={}& \bm{\mu}^- + K\bm{v} \label{E:kf_mean} \\
  P^+ ={}& P^- - KHP^-\textit{,} \label{E:kf_cov}
\end{align}
where $K = P^-H^\top S^{-1}$ is the Kalman gain,
and $\bm{v} = \bm{z} - H\bm{\mu}^-$ and $S = HP^-H^\top + R$ are the measurement and covariance innovations.
$R$ is a diagonal $m~\times~m$ matrix of altitude-dependent variances $\sigma^2_{s,i}$ associated with each measurement $z_i$,
and $H$ is a $m~\times~n$ matrix denoting a linear sensor model
that intrinsically selects part of the state $\{\zeta_1, \ldots, \zeta_m\}$ observed through $\bm{z}$.
The information to account for variable-resolution measurements is incorporated in a simple manner through the sensor model $H$
as detailed in the following section.

The constant-time updates in Equations~\eqref{E:kf_mean} and~\eqref{E:kf_cov} are repeated every time new data is registered.
Note that, as all models are linear in this case, the \ac{KF} update produces the optimal solution.
This approach also permits fusing heterogeneous sensor information into a single map.

\subsection{Altitude-dependent Sensor Model} \label{s:sensor_model}

We consider that the information collected by an on-board camera degrades with altitude in two ways: (i)~noise and (ii)~resolution.
The proposed sensor model accounts for these issues in a probabilistic manner as follows.

\begin{SCfigure}[][!h]
  \includegraphics[width=0.23\textwidth]{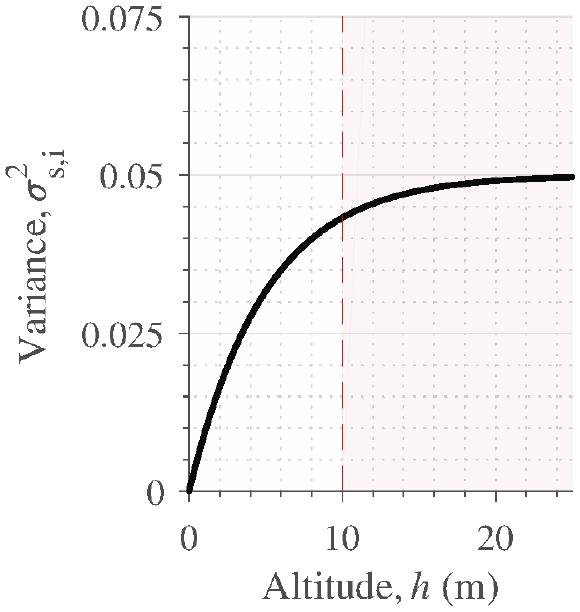}
   \caption{Sensor noise model for a snapshot camera providing measurements as $\mathcal{N}(\mu_{s,i},
   \sigma_{s,i})$ with $a = 0.2$, $b = 0.05$ in~\eqnref{E:sensor_model}.
   The uncertainty $\sigma_{s,i}^2$ increases with $h$ to represent degrading image quality.
   The dotted line at $h = 10$\,m indicates the altitude above which image resolution scales down by a factor of 2.
   }\label{F:sensor_model}
\end{SCfigure}

We assume an altitude-dependent Gaussian sensor noise model.
For each observed point $\textbf{x}_i \in X$, the camera provides a measurement $z_i$ capturing the target field $\zeta_i$ as
$\mathcal{N}(\mu_{s,i}, \sigma_{s,i})$, where $\sigma_{s,i}^2$ is the noise variance expressing uncertainty in $z_i$.
To account for lower-quality images taken with larger camera footprints,
$\sigma_{s,i}^2$ is modelled as increasing with \ac{UAV} altitude $h$ using:
\begin{align}
 \sigma_{s,i}^2 ={}& a(1-e^{-bh})\textit{,} \label{E:sensor_model}
\end{align}
where $a$ and $b$ are positive constants.
As an example, \figref{F:sensor_model} shows the sensor noise model used in our agricultural monitoring set-up, which represents a snapshot camera.
The measurements $z_i$ denote green biomass level computed from calibrated spectral indices in the images~\citep{Liebisch2017}.

Moreover, we define altitude envelopes corresponding to different image resolution scales with respect to the initial points $X$.
At higher altitudes and lower resolutions, adjacent $\textbf{x}_i$ are indexed by a single sensor measurement $z_i$ through the sensor model $H$.
At the maximum mapping resolution, $H$ is simply used to select the part of the state observed with a scale of $1$.
However, to handle lower-resolution data, the elements of $H$ are used to map multiple $\zeta_i$ to a single $z_i$
scaled by the square inverse of the resolution scaling factor $s_f$.
Note that the fusion described in~\secref{s:data_fusion} is always performed at the maximum mapping resolution,
so the proposed model $H$ considers low-resolution measurements as an scaled average of the high-resolution map.

\section{PLANNING APPROACH} \label{S:planning_approach}
This section summarizes our planning strategy,
which generates fixed-horizon plans through a combination of a 3-D grid search and evolutionary optimization.
We describe our approaches to parametrizing trajectories and computing the informative objective with the new map representation.
For further details, the reader is referred to our previous work~\citep{Popovic2017}.

\subsection{Trajectory Parametrization} \label{s:trajectory_parametrization}
We parametrize a polynomial trajectory $\psi$ with a sequence of $N$ control waypoints to visit 
$\mathcal{C} = [\textbf{c}_1, \ldots, \textbf{c}_N]$ connected using $N-1$ $k$-order spline segments for minimum-snap dynamics~\citep{Richter2013}.
The first waypoint $\textbf{c}_1$ is clamped as the initial \ac{UAV} position.
As discussed in~\secref{S:problem_statement}, the function \textsc{measure(\textperiodcentered)} in~\eqnref{E:ipp_problem} is defined by computing the spacing of measurement sites
along $\psi$ given a constant sensor frequency.

\subsection{Planning} \label{s:planning}

We create adaptive plans using a fixed-horizon approach,
alternating between replanning and execution until the elapsed time $t$ exceeds the budget $B$.
Each new plan is a polynomial defined by $N$ control waypoints.
Our replanning procedure (\algref{A:replan_path})
involves obtaining an initial trajectory through a 3-D grid search (Lines~3-6),
and then refining it using evolutionary optimization (Line~7), as detailed below.

To evaluate the utility $I$ in~\eqnref{E:ipp_problem} for a point $\textbf{c}$, we maximize uncertainty reduction in the map,
measured by the covariance $P$ as:
\begin{align}
 I[\textbf{c}] ={}& \Tr(P^-) - \Tr(P^+) \textit{,} \label{E:info_objective}
\end{align}
where $\Tr($\textperiodcentered$)$ is the trace of a matrix,
and the superscripts on $P$ denote the prior and posterior covariances as evaluated by~\eqnref{E:kf_cov}.
Note that, while~\eqnref{E:info_objective} defines $I$ for a single measurement,
we use the same principles to determine the utility of a trajectory
by fusing a sequence of measurements and computing the overall reduction in $\Tr(P)$.
\begin{algorithm}[!h]
\renewcommand{\algorithmicrequire}{\textbf{Input:}}
\renewcommand{\algorithmicensure}{\textbf{Output:}}
\algrenewcommand\algorithmiccomment[2][\scriptsize]{{#1\hfill\(\triangleright\)
\textcolor[rgb]{0.4, 0.4, 0.4}{#2} }}
\begin{algorithmic}[1]

  \Require Covariance matrix of current model $P$, number of control waypoints $N$, lattice points $\mathcal{L}$
  \Ensure Waypoints defining next polynomial plan $\mathcal{C}$
  
  \State $P' \gets P$ \Comment{Create local copy of covariance matrix.}
  \State $\mathcal{C} \gets \emptyset$ \Comment{Initialize control points.}
  \While {$N \geq |\mathcal{C}|$}
  \State $\textbf{c}^* \gets$ Select viewpoint in $\mathcal{L}$ using~\eqnref{E:ipp_problem}
  \State $P' \gets$ \Call{update\_cov}{$P'$, $\textbf{c}^*$} \Comment{Using Bayesian fusion.}
  \State $\mathcal{C} \gets \mathcal{C} \cup \textbf{c}^*$
  \EndWhile
  \State $\mathcal{C} \gets$ \Call{cmaes}{$\mathcal{C}$, $P$}
\Comment{Optimize polynomial.}

\end{algorithmic}
\caption{\textsc{replan\_path} procedure}\label{A:replan_path}
\end{algorithm}

A value-dependent objective is defined
to focus on higher-valued regions of interest requiring infestation treatment.
We formalize this in an adaptive \ac{IPP} setting using a low threshold $\mu_{th}$
to seperate the interesting (above, possibly infected) and uninteresting (below, likely uninfected) biomass value range.
Hence, \eqnref{E:info_objective} is modified
so that elements of $\Tr(P)$ mapping to the mean of each cell $\mu_i < \mu_{th}$ via location $\textbf{x}_i \in X$
are excluded from the objective computation.

\subsubsection{3-D Grid Search} \label{sss:grid_search}
\begin{SCfigure}[][h]
  \includegraphics[width=0.258\textwidth]{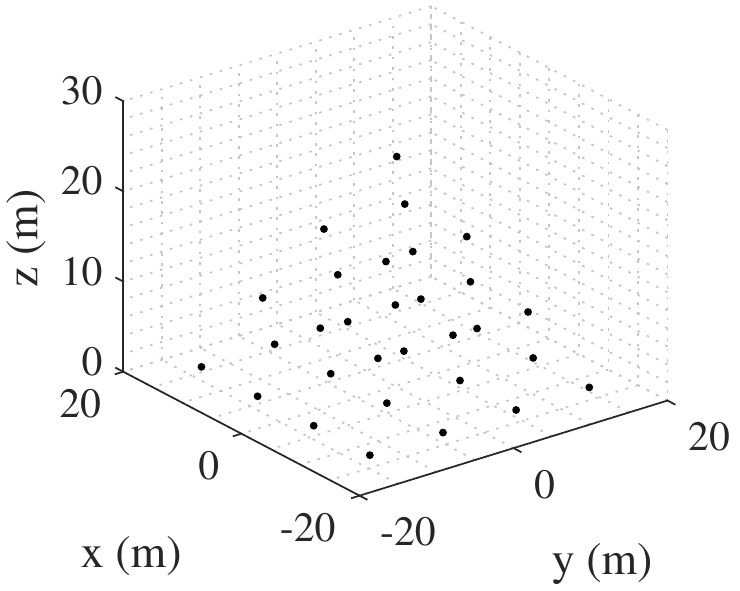}
   \caption{A visualization of the 3-D lattice grid $\mathcal{L}$ used in our simulation experiments (30 points).
    The length scales are defined to efficiently obtain an initial trajectory solution for the problem
    given the computational resources.}\label{F:lattice}
\end{SCfigure}

The first replanning step (Lines~3-6) supplies an initial solution for optimization in~\secref{sss:optimization}.
To achieve this, a 3-D grid search is performed based on a coarse multiresolution lattice $\mathcal{L}$
in the \ac{UAV} configuration space (\figref{F:lattice}).
We quickly obtain a low-accuracy solution neglecting sensor dynamics by using
the points $\mathcal{L}$ to represent candidate measurement sites and assuming constant velocity travel.
In this set-up, we conduct a sequential greedy search for $N$ waypoints (Line~3).
The next best point $\textbf{c}^*$ (Line~4)
is found by evaluating~\eqnref{E:ipp_problem} with the utility $I$ in~\eqnref{E:info_objective} over $\mathcal{L}$.
For each $\textbf{c}^*$, we simulate a fused measurement via~\eqnref{E:kf_cov} (Line~5),
and add it to the initial trajectory solution (Line~6).

\subsubsection{Optimization} \label{sss:optimization}
The second replanning step (Line~7) optimizes the grid search solution by solving~\eqnref{E:info_objective}
in~\secref{sss:grid_search} for a sequence of measurements taken along the trajectory.
For global optimization,
we propose the Covariance Matrix Adaptation Evolution Strategy (CMA-ES)~\citep{Hansen2006}.
This choice is motivated by the nonlinearity of the objective space (\eqnref{E:ipp_problem})
as well as previous results~\citep{Popovic2017,Hitz2014}.
In~\secref{s:optimization_comparison}, we compare our proposed optimizer to alternatives and demonstrate its values.

\section{EXPERIMENTAL RESULTS} \label{S:experimental_results}
In this section, we validate our \ac{IPP} framework in simulation by comparing it to state-of-the-art methods
and study the effects of using different optimization routines in our algorithm.
We then show proof of concept by using our system to map color in different indoor environments.

\subsection{Comparison Against Benchmarks} \label{s:benchmark_comparison}
Our framework is evaluated on $30$ $30\times30$\,m simulated Gaussian random field environments with cluster radii ranging from $1$\,m to $3$\,m.
We use a uniform resolution of $0.75$\,m for both the training $X$ and predictive $X^*$ grids,
and perform uninformed initialization with a mean prior of $50$\,\% green biomass.
For the \ac{GP}, an isotropic Mat\'ern 3/2 kernel (\eqnref{E:matern_kernel}) is applied
with hyperparameters $\bm{\theta} = \{\sigma_n^2,\,\sigma_f^2,\,l\} = \{1.42,\,1.82,\,3.67\}$
trained from $4$ independent maps with the variances modified
to cover the full $0$ to $100$\% green biomass range during inference.

For fusion, measurement noise is simulated based on the camera model in~\figref{F:sensor_model},
with a $10$\,m altitude beyond which images scale by a factor of $s_f = 0.5$.
This places a realistic limit on the quality of data that can be obtained from higher altitudes.
We set a square camera footprint with $60\degree$ \ac{FoV} and a $0.15$\,Hz measurement frequency.

Our approach is compared against traditional ``lawnmower'' coverage
and the sampling-based rapidly exploring information gathering tree (RIG-tree)
introduced by~\citet{Hollinger2014}, a state-of-the-art \ac{IPP} algorithm.
A $200$\,s budget $B$ is specified for all three methods.
To evaluate performance, we quantify uncertainty with the covariance matrix trace $\Tr(P)$
and consider the \ac{RMSE} and \ac{MLL} at points in $X$ with respect to ground truth as accuracy statistics.
As described by~\citet{Marchant2014}, the \ac{MLL} is a probabilistic confidence measure which
incorporates the variance of the predictive distribution.
Intuitively, all metrics are expected to reduce over time as data are acquired,
with steeper declines signifying better performance.

We specify the initial \ac{UAV} position as ($7.5$\,m, $7.5$\,m) within the field with $8.66$\,m altitude.
For trajectory optimization, the reference velocity and acceleration are $5$\,m$/$s and $2$\,m$/$s$^2$
using polynomials of order $k = 12$,
and the number of measurements along a path is limited to $10$ for computational feasibility.
In our planner, we define polynomials by $N = 5$ waypoints and use the lattice in~\figref{F:lattice} for the 3-D grid search.
In RIG-tree, we associate control waypoints with vertices,
and form polynomials by tracing the parents of leaf vertices to the root.
For both planners, we consider the utility $I$ in~\eqnref{E:info_objective} and set a threshold of $\mu_{th} = 40\%$.

As outlined in our previous paper~\citep{Popovic2017}, we use an adaptive version of RIG-tree
which alternates between tree construction and plan execution.
The branch expansion step size is set to $10$\,m for best performance based on multiple trials.

In the coverage planner, height ($8.66$\,m) and velocity ($0.78$\,m$/$s) are defined
for complete coverage given the specified budget and measurement frequency.
To design a fair benchmark, we studied possible ``lawnmower'' patterns
while changing velocity to match the budget, then selected the best-performing one.

\begin{figure}[!h]
\centering
  \includegraphics[width=0.68\columnwidth]{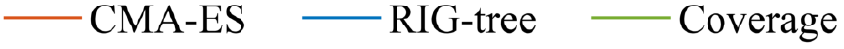}
  \includegraphics[width=\columnwidth]{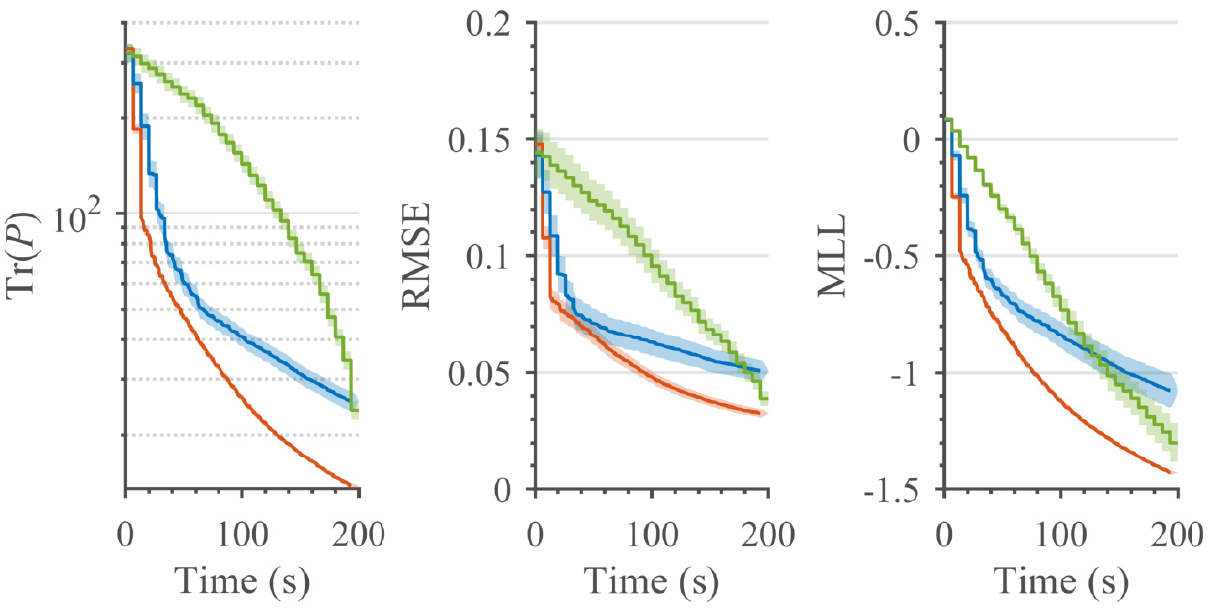}
   \caption{Comparison our \ac{IPP} algorithm using the CMA-ES against the RIG-tree and coverage planners.
   The solid lines represent means over 30 trials. The thin shaded regions depict 95\% confidence bounds.
   Using \ac{IPP}, uncertainty (left) and error (middle, right) reduce quickly
   as the \ac{UAV} obtains low-resolution images before descending. Note that $\Tr(P)$ is plotted on a logarithmic scale.}\label{F:methods}
\end{figure}

\figref{F:methods} shows how the metrics evolve for each planner during the mission.
For our algorithm, we use the proposed CMA-ES optimization method.
The coverage curve (green) validates our previous result~\citep{Popovic2017}
that uncertainty (left) reduces uniformly for a constant altitude and velocity.
This motivates \ac{IPP} approaches, which perform better because they compromise between resolution and \ac{FoV},
as illustrated in~\figref{F:teaser}.

Our algorithm (red) produces maps with lower uncertainty and error
than those of RIG-tree (blue) given the same budget.
This confirms that our two-stage planner is more effective than sampling-based methods
with the new mapping strategy.
We noted that fixed step-size is a key drawback of RIG-tree,
because high values allowing initial ascents tend to limit incremental navigation when later refining the map.

\begin{figure*}[!h]
\centering
\begin{subfigure}{.48\columnwidth}
\includegraphics[width=\columnwidth]{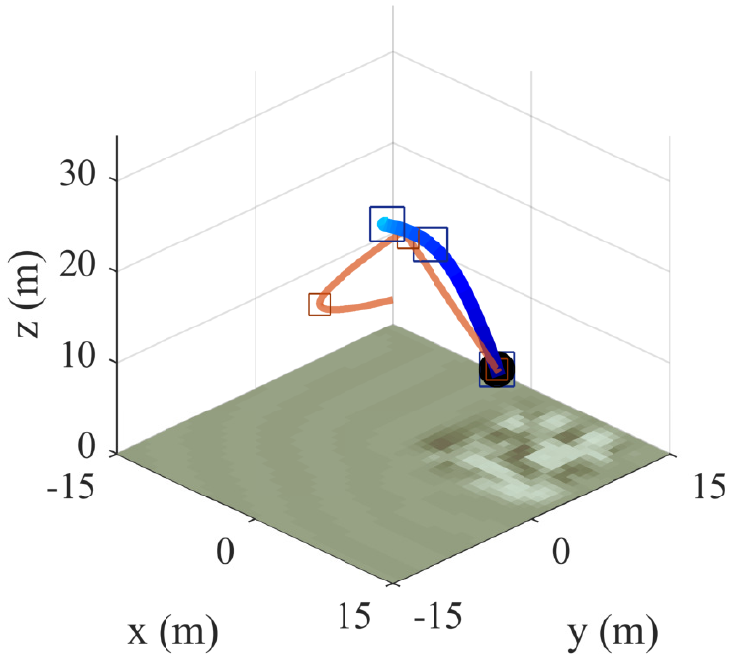}%
\end{subfigure}\hfill%
\begin{subfigure}{.48\columnwidth}
\includegraphics[width=\columnwidth]{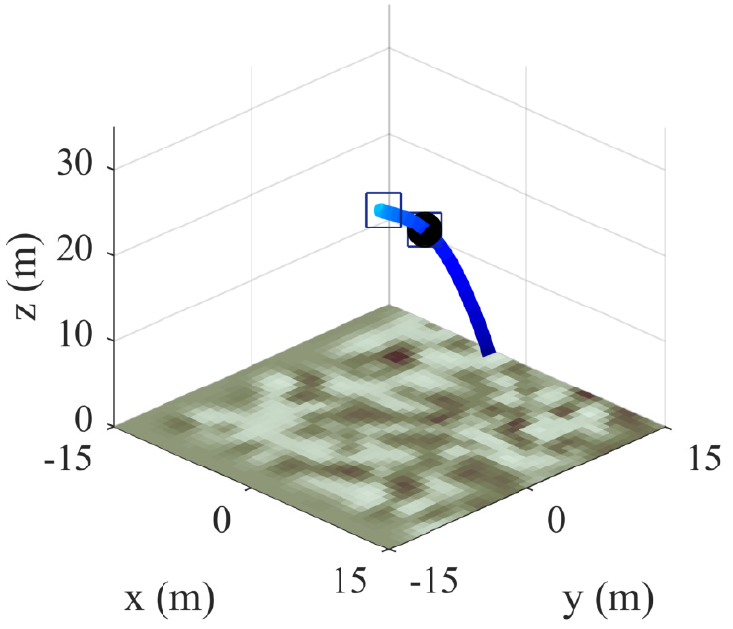}%
\end{subfigure}\hfill%
\begin{subfigure}{.48\columnwidth}
\includegraphics[width=\columnwidth]{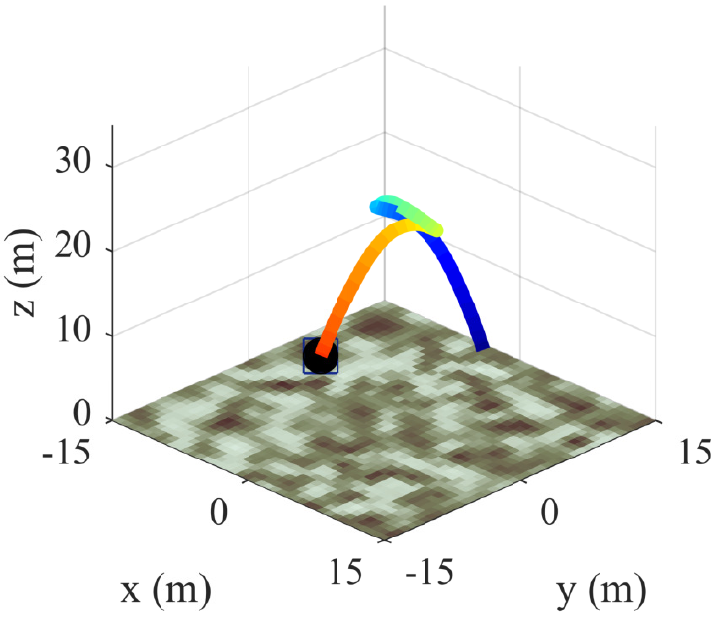}%
\end{subfigure}\hfill%
\begin{subfigure}{.48\columnwidth}
\centering
\includegraphics[width=\columnwidth]{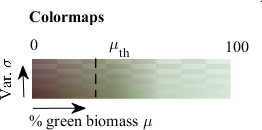} \\
\vspace{1.5mm}
\includegraphics[width=0.8\columnwidth]{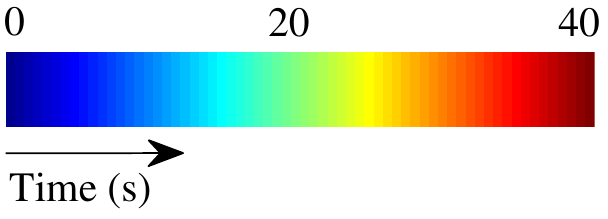}%
\end{subfigure}%
\vspace*{1.5mm}
\hspace*{2mm}
\begin{subfigure}[]{.33\columnwidth}
\includegraphics[width=\columnwidth, right]{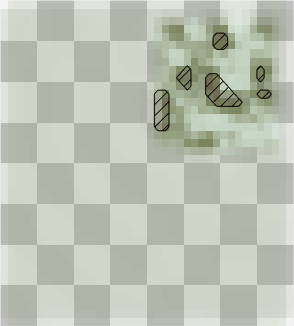}%
\end{subfigure}
\hspace*{15mm}
\begin{subfigure}{.33\columnwidth}
\includegraphics[width=\columnwidth]{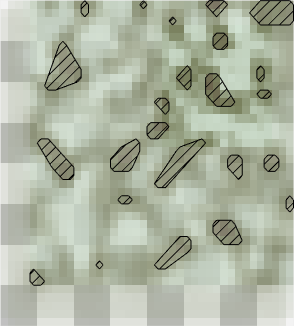}%
\end{subfigure}
\hspace*{15mm}
\begin{subfigure}{.33\columnwidth}
\includegraphics[width=\columnwidth]{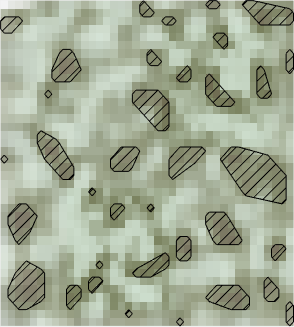}%
\end{subfigure}
\hspace{6mm}
\begin{subfigure}{.40\columnwidth}
\includegraphics[width=\columnwidth]{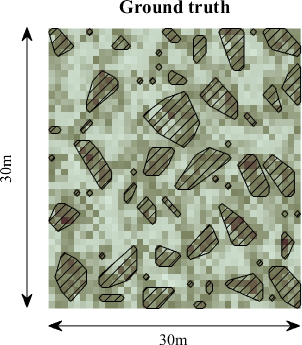}%
\end{subfigure}%
\caption{Example simulation results of our \ac{IPP} framework. The colorbars are shown on the top-right.
Greener and browner shades represent high and low values of the target biomass parameter, respectively.
In the bottom maps, opacity indicates the uncertainty (variance, $\sigma_i^2$) of the model,
with the checkerboard added for visual clarity and the hatched sections denoting uninteresting areas with $< \mu_{th} = 40\%$.
The simulated ground truth is shown in the bottom-right.
The three columns on the left depict the trajectories (top row) and maps (bottom row) at
different snapshots of the mission at times $t = 0$\,s, $6.67$\,s, and $33.74$\,s.
In the top plots, the black dot indicates the current \ac{UAV} position while the squares show the measurement sites.
The top-left figure illustrates an example trajectory before (orange) and after (colored gradient) optimization using the CMA-ES.
Note that the map means are rendered in the top trajectory plots.
}
\label{F:map_evolution}
\end{figure*}

\figref{F:map_evolution} depicts the progression of our \ac{IPP} framework for a measurement sequence
in an example simulation trial.
The top and bottom row visualize the planned \ac{UAV} trajectories and maps,
respectively, as images are fused.
The top-left plot depicts the first planned trajectory before (orange) and after (colored gradient) using the CMA-ES.
As shown, applying optimization shifts initial measurement sites (squares) to high altitudes,
allowing low-resolution data to be collected quickly before refining the map (second and third columns).
A visual ground-truth comparison (bottom-right) confirms that
our method produces a fairly complete map with most uninteresting regions (hatched areas) identified.

\subsection{Optimization Method Comparison} \label{s:optimization_comparison}
Next, we examine the effects of using different optimization routines on the 3-D grid search output in~\secref{sss:optimization}
to evaluate our proposed CMA-ES approach.
We consider the following methods in the same simulation set-up as above:
\begin{itemize}
 \item \textit{Lattice}: grid search only (i.e. without Line~7 in~\algref{A:replan_path}),
 \item \textit{CMA-ES}: global evolutionary optimization (as in~\secref{sss:optimization}),
 \item \textit{\ac{IP}}: approximate gradient-based optimization using interior-point approach~\citep{Byrd2006}, 
 \item \textit{\ac{BO}}: global optimization using a \ac{GP} process model~\citep{Gelbart2014}.
\end{itemize}

We allocate approximately the same amount of optimization time for each method.
For the CMA-ES, we set step-sizes of $3$\,m and $4$\,m in the planar and vertical (altitude) co-ordinate directions, respectively.
For the local \ac{IP} optimizer, we approximate Hessians by a dense quasi-Newton strategy
and apply the step-wise algorithm described by~\citet{Byrd2006}.
For \ac{BO}, we use a time-weighted Expected Improvement acquisiton function studied by~\citet{Gelbart2014}.

\tabref{T:methods} displays the mean results for each method averaged over $30$ trials, with the benchmarks included for reference.
Following~\citet{Marchant2014}, we also show weighted statistics to emphasize errors in high-valued regions.
As the same objective is used for all methods,
consistent trends are observed in both non-weighted and weighted metrics.

Comparing the lattice approach with the CMA-ES and \ac{IP} methods confirms that optimization reduces both uncertainty and error.
With lowest values, the proposed CMA-ES performs best on all indicators as it searches globally to escape local minima.
This effect is evidenced in~\figref{F:map_evolution}, where early measurements are moved to higher altitudes.
Surprisingly, however, applying \ac{BO} yields mean metrics poorer than those of the lattice.
From inspection, this is likely due to its high exploratory behaviour
causing erratic paths and hence worse performance at later planning stages.
We faced similar issues when using the CMA-ES with large step-sizes.
\begin{table}[h]
\resizebox{\columnwidth}{!}{%
\begin{tabular}{lSSSSS} \toprule
    Method         & {$\Tr(P)$}      & {RMSE}    & {WRMSE}    & {MLL}      & {WMLL}    \\ \midrule \midrule
    Lattice        & 56.193          & 0.0624    & 0.0622     & -0.880    & -0.881   \\ \midrule
    CMA-ES         & 46.780          & 0.0541    & 0.0536     & -0.976    & -0.981   \\ \midrule
    IP             & 51.628          & 0.0575    & 0.0574     & -0.918    & -0.919   \\ \midrule
    BO             & 62.121          & 0.0646    & 0.0642     & -0.805    & -0.808   \\ \midrule
    RIG-tree       & 68.581          & 0.0696    & 0.0696     & -0.755    & -0.757   \\ \midrule
    Coverage       & 165.121         & 0.0972    & 0.0972     & -0.685    & -0.688   \\ \bottomrule
\end{tabular}
}
\caption{Mean informative metrics for all algorithms, averaged over $30$ trials.
The lowest uncertainties and errors obtained with the CMA-ES justify our proposed global optimization approach.} \label{T:methods}
\end{table}
\subsection{Experiments} \label{s:experiments}
\begin{figure*}[!h]
\centering
\begin{subfigure}{.42\columnwidth}
\includegraphics[width=\columnwidth]{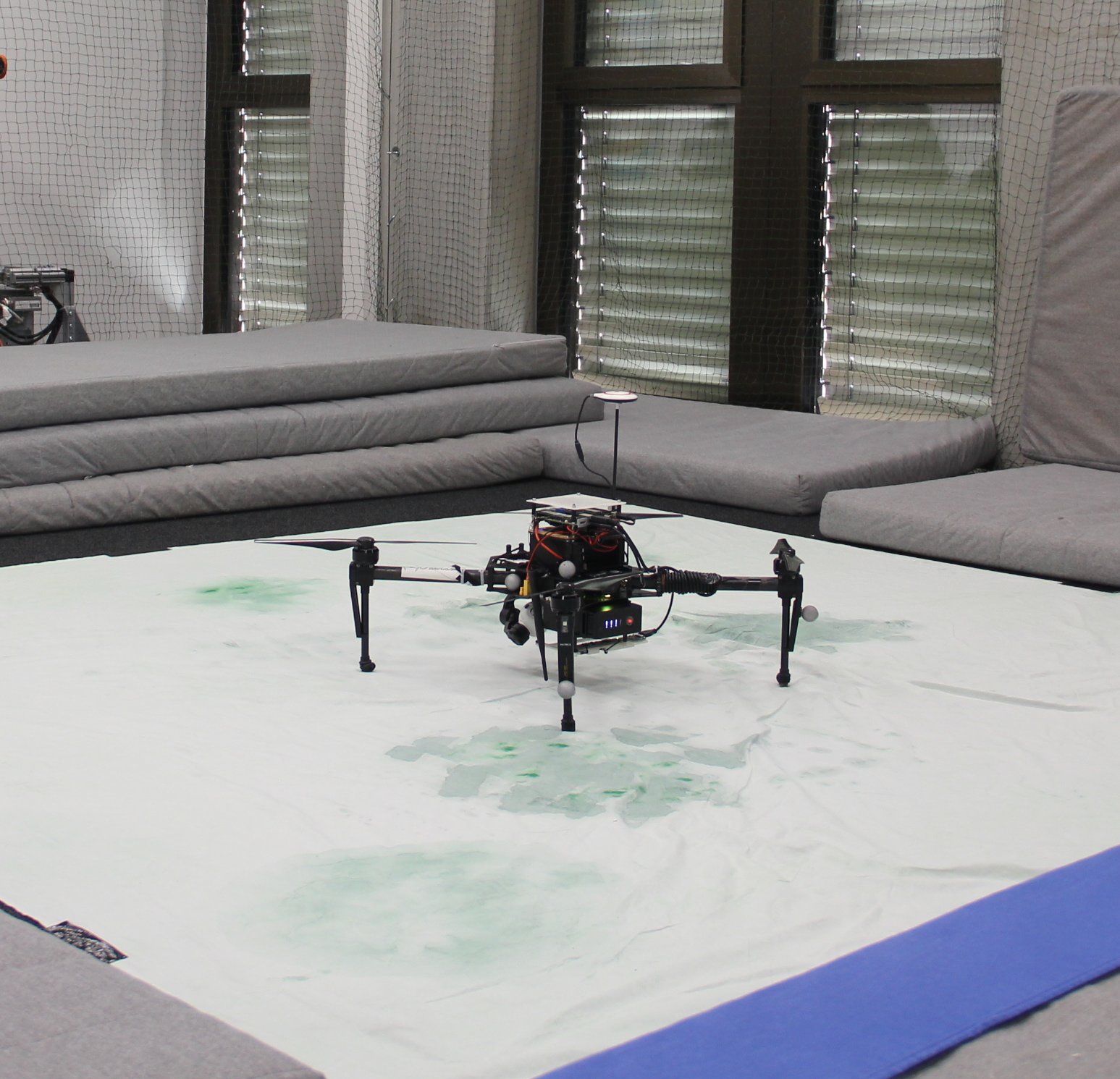}%
\caption{}%
\label{SF:uav}%
\end{subfigure}\hfill%
\begin{subfigure}{.42\columnwidth}
\includegraphics[width=\columnwidth]{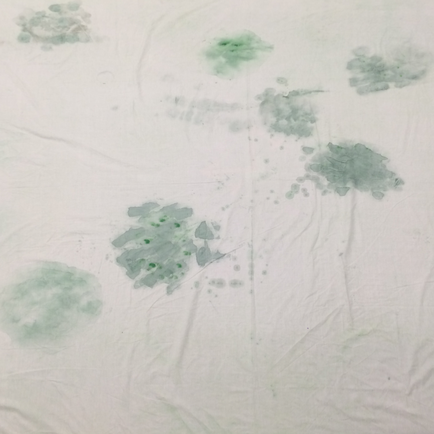}%
\caption{}%
\label{SF:sheet1}%
\end{subfigure}\hfill%
\begin{subfigure}{.42\columnwidth}
\includegraphics[width=\columnwidth]{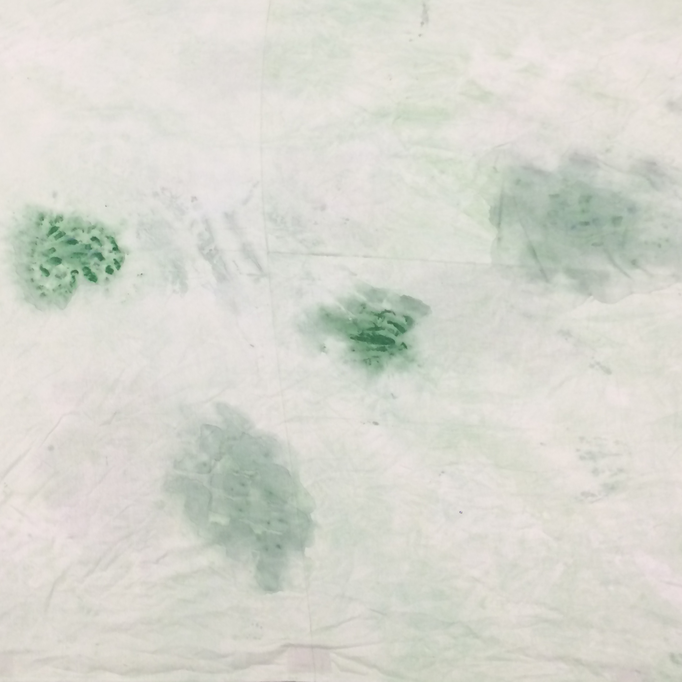}%
\caption{}%
\label{SF:sheet2}%
\end{subfigure}\hfill%
\begin{subfigure}{.42\columnwidth}
\includegraphics[width=\columnwidth]{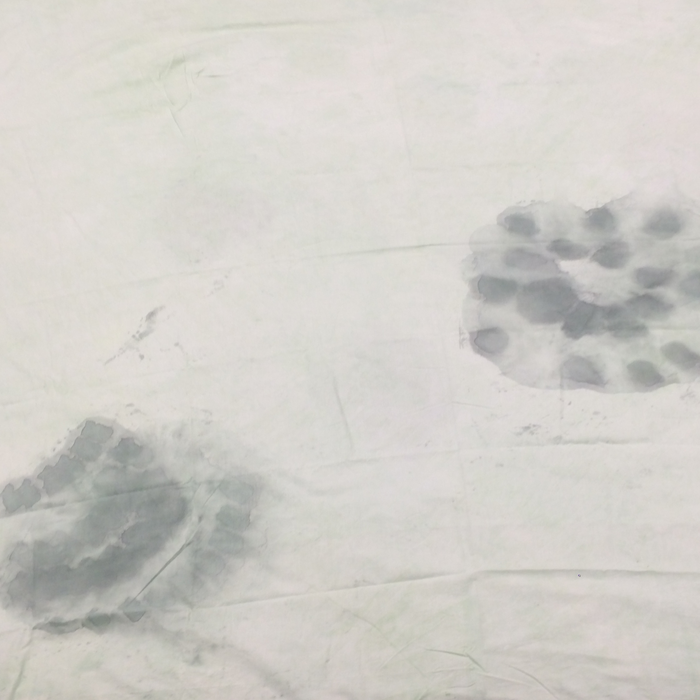}%
\caption{}%
\label{SF:sheet3}%
\end{subfigure}\hfill%
\caption{(a) shows a side view of our experimental set-up.
The sheets in (b)-(d) simulate vegetation distributions for saturation mapping
using colored pointclouds from the on-board depth camera.
The painted regions correspond to more highly saturated areas.}\label{F:exp_setup}
\end{figure*}

We show our \ac{IPP} strategy running in real-time on a DJI Matrice 100~\citep{Sa2017}.
The experiments are conducted in an empty $2\times2$\,m indoor environment with a maximum altitude of $2$\,m
and the Vicon motion capture system for state estimation (\figref{SF:uav}).
To establish the potential of our approach in agriculture, we mimic vegetation detection
by mapping the normalized saturation level of painted green sheets (Figs. \ref{SF:sheet1}-\ref{SF:sheet3}).
A downward-facing Intel RealSense ZR300 depth camera provides colored pointcloud measurements.

We set a $0.1$\,m resolution grid for both \ac{GP} training and prediction,
and follow the method in~\secref{s:benchmark_comparison} to train the model
with the same Mat\'ern kernel.
To take measurements, the saturation values received are first averaged per cell.
Then, Bayesian fusion is performed using the sensor model in~\eqnref{E:sensor_model} with $a = 0.1$, $b = 0.2$,
and a $1.5$\,m altitude above which images scale by $s_f = 0.5$.
We opt for a depth camera to obtain sufficient noise variation within the altitude range,
noting that our sensing and localization interfaces are adaptable for field trials.

\begin{figure}[!h]
\centering
\begin{subfigure}{\columnwidth}
\includegraphics[width=0.54\columnwidth]{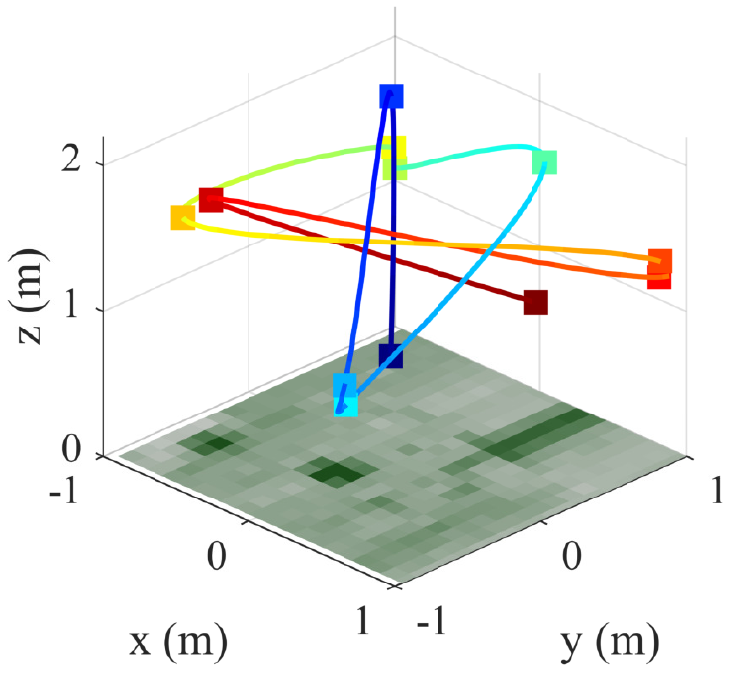}%
\hfill
\raisebox{4.5mm}{
\includegraphics[width=0.34\columnwidth]{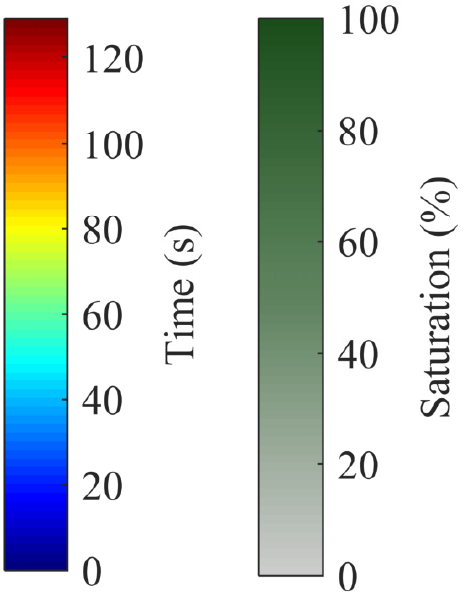}}%
\caption{}%
\label{SF:exp_path}%
\end{subfigure} \\
\vspace{1mm}
\begin{subfigure}{0.985\columnwidth}
\includegraphics[width=\columnwidth]{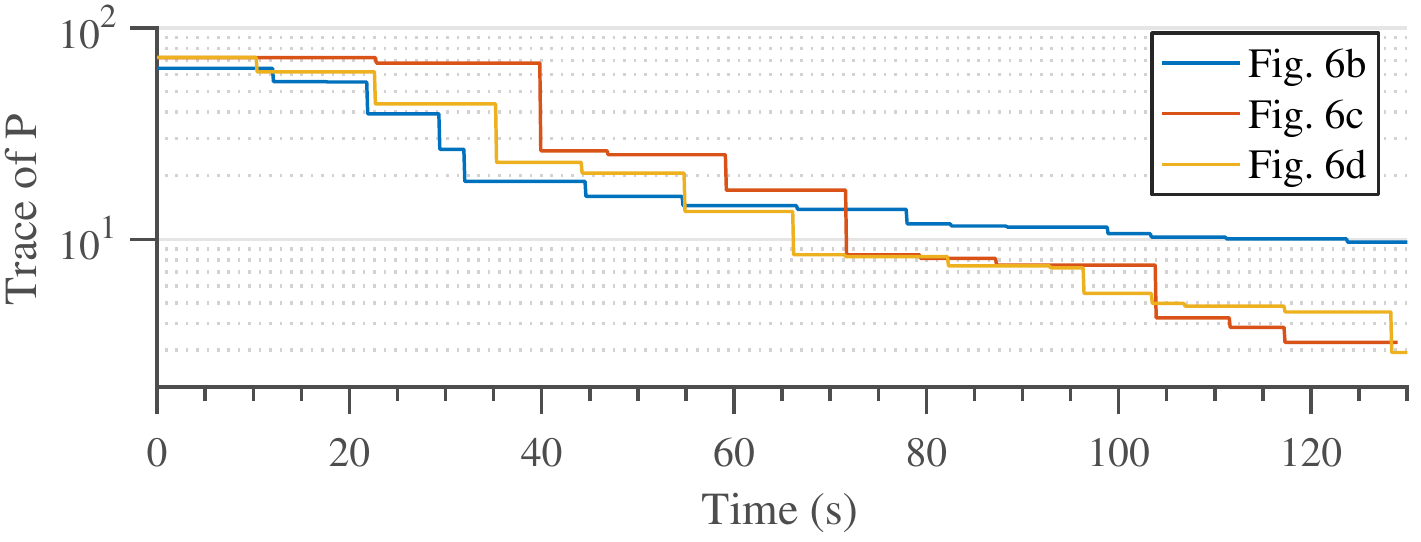}%
\caption{}%
\label{SF:exp_uncertainty}%
\end{subfigure}
\caption{(a) depicts the recorded trajectory for our \ac{IPP} algorithm mapping the sheet in~\figref{SF:sheet2}.
The squares indicate measurement sites.
The darker cells in the rendered final map represent successfully detected areas with high paint saturation.
The curves in (b) show uncertainty variations for the different sheets,
thus validating our approach.
Note that planning time is taken into account.}\label{F:exp_results}
\end{figure}

The aim is to show our framework mapping different continuous and realistic scalar fields.
We consider the three target distributions shown in~\figref{F:exp_setup}.
In each case, the initial measurement point is set to $0.8$\,m at the map center.
We specify the planning budget $B$ as $130$\,s, and
define polynomials by $N\,=\,4$ control waypoints
with a reference velocity and acceleration of $1.5$\,m$/$s and $2$\,m$/$s$^2$,
and a $0.2$\,Hz camera frequency.
The utility $I$ is quantified by~\eqnref{E:info_objective} as before, with $\mu_{th} = 20\%$
and the CMA-ES used for optimization.

\figref{F:exp_results} summarizes our experimental results.
As an example, \figref{SF:exp_path} visualizes the traveled trajectory for the sheet in~\figref{SF:sheet2}.
The darker map cells correspond to detected painted areas.
As in~\figref{F:map_evolution}, the \ac{UAV} remains at high-resolution altitudes upon initially ascending.
Note that this effect is less evident due to the narrow camera \ac{FoV} limiting information gain.
In~\figref{SF:exp_uncertainty}, the uncertainty reductions in each scenario (solid curve)
validate the applicability of our approach to different environments in real-time.
Future work will study increasing efficiency
using submaps~\citep{Sun2015} or other approximation methods~\citep{Rasmussen2006}.

\section{CONCLUSIONS AND FUTURE WORK} \label{S:conclusions}
This work introduced a mapping approach for \ac{IPP} in terrain monitoring using \acp{UAV}.
Our method uses the spatial correlation encoded in \ac{GP} models to generate prior maps
for recursive filtering updates fusing variable-resolution data from probabilistic sensors.
The resulting maps are employed for \ac{IPP}
by using an evolutionary technique to optimize trajectories initialized by a 3-D grid search.

We evaluated our framework on the application of agricultural biomass monitoring.
In simulation, we showed its advantages over state-of-the-art planners and alternative optimization schemes
in terms of informative metrics.
Proof of concept experiments validated the applicability of our approach in different scenarios
with real-time requirements.

Future work will address improving efficiency for larger-scale field trials.
Interesting research directions involve capturing temporal, as well as spatial, correlations,
and incorporating prior knowledge from previous scans.

\section*{ACKNOWLEDGMENT}
This project has received funding from the European Union’s Horizon 2020 
research and innovation programme under grant agreement No 644227 and from the 
Swiss State Secretariat for Education, Research and Innovation (SERI) under 
contract number 15.0029.
We would like to thank Dr. Frank Liebisch for his useful discussions.

\bibliographystyle{IEEEtranN}
\footnotesize
\bibliography{references/2017-iros-popovic}

\end{document}